\documentclass{article}

\usepackage{arxiv}

\usepackage[utf8]{inputenc}
\usepackage[T1]{fontenc}
\usepackage{hyperref}
\hypersetup{hidelinks}
\usepackage{url}
\usepackage{booktabs}
\usepackage{amsfonts}
\usepackage{nicefrac}
\usepackage{microtype}
\usepackage{lipsum}
\usepackage{graphicx}
\usepackage{amsmath, amssymb, amsfonts}
\usepackage{amsthm}

\usepackage{subcaption}

\usepackage{rotating}            
\usepackage{multirow}
\usepackage{array}
\usepackage{ragged2e}
\graphicspath{ {./images/} }

\title{TeleHairing: A Teleoperation Baseline for Robotic Haircutting}

\author{
\begin{minipage}{0.96\textwidth}
\centering
\textbf{Zhendai Huang\textsuperscript{1}, Aleksi Vilkki\textsuperscript{1}, Jianan Huang\textsuperscript{2,3}, Bolin Liao\textsuperscript{4,5},}\\[3pt]
\textbf{Chunbo Luo\textsuperscript{6}, Leopoldo Angrisani\textsuperscript{7}, Shuai Li\textsuperscript{1,8,*}}\\[8pt]
\normalfont\small
\textsuperscript{1}Faculty of Information Technology and Electrical Engineering, University of Oulu,\\
90570 Oulu, Finland\\[2pt]
\textsuperscript{2}Research Unit of Health Sciences and Technology (HST), Faculty of Medicine,\\
University of Oulu, 90220 Oulu, Finland\\[2pt]
\textsuperscript{3}Research Unit of Disease Network, Faculty of Biochemistry and Molecular Medicine,\\
University of Oulu, 90220 Oulu, Finland\\[2pt]
\textsuperscript{4}School of Information Science and Engineering, Provincial Key Laboratory of Informational\\
Service for Rural Area of Southwestern Hunan, Shaoyang University, 422000 Shaoyang, China\\[2pt]
\textsuperscript{5}College of Computer Science and Engineering, Jishou University, 416000 Jishou, China\\[2pt]
\textsuperscript{6}Department of Computer Science, University of Exeter, Exeter, United Kingdom\\[2pt]
\textsuperscript{7}Department of Information Technology and Electrical Engineering,\\
University of Naples Federico II, Naples, Italy\\[2pt]
\textsuperscript{8}VTT Technical Research Centre of Finland, 90570 Oulu, Finland\\[6pt]
\textsuperscript{*}Corresponding author: \href{mailto:shuai.li@oulu.fi}{\texttt{shuai.li@oulu.fi}}
\end{minipage}
}

\begin{document}
\maketitle
\begin{abstract}
Robotic haircutting requires controlled tool motion near the head while simultaneously accounting for communication, visual feedback, tool actuation, and interruption handling. Existing studies still lack an operator-in-the-loop reference for analyzing these coupled behaviors before human trials or stronger autonomy. This paper presents TeleHairing, a closed-loop teleoperation architecture for mannequin-based robotic haircutting evaluation under local, relay, and remote deployment conditions. Logged timing shows that the main remote latency increase occurs before the robot-side control endpoint: overall timing reached 190.5~ms in remote mode, while robot-side command queue, control processing, and control-to-robot timing remained similar across modes. Trajectory analysis shows that the larger remote command-following error was dominated by the terminal withdrawal segment rather than accumulated uniformly over the path; excluding this segment reduced remote root-mean-square error (RMSE) from 45.1~mm to 9.6~mm. Detection-loss trials further show that rebase events resumed motion without a large target jump under the tested condition. These results clarify how deployment, execution, and interruption affect the robotic haircutting teleoperation loop, providing a quantitative reference for future autonomy, safety, and user-facing studies.
\end{abstract}


\section{Introduction}\label{sec.1}
Haircutting is a personal-care task that requires close physical interaction and skilled manual control. It also requires continuous adaptation to an individual's head shape, hair condition, and styling intent. Interest in alternatives to conventional in-person haircutting has increased when access to human barbers is limited, such as during the COVID-19 pandemic and in assistive personal-care settings~\cite{yougov2020homehaircutus}. Related robotic studies on head-proximate grooming, including combing and compliant hair manipulation, suggest that near-head hair interaction is feasible. They also show that the task remains contact-rich, comfort-sensitive, and operationally constrained~\cite{dennler2021design,yoo2025softcompliant}. Robotic haircutting is distinct from generic manipulation because tool motion must be coordinated near the head while hair geometry, visual occlusion, tool actuation, and user-comfort constraints remain part of the operating context. These observations motivate robotic investigation of haircutting.

At the same time, autonomous robotic haircutting remains a challenging and not yet well-defined problem. The robot must move cutting tools near sensitive regions while maintaining continuous spatial control under limited safety margins~\cite{khan2026haircuttingsurvey}. The task is also affected by variation in head geometry, hair density, hair texture, hair length, and user preference~\cite{li2025theorypractice}. Recent reviews identify real-time decision-making, hair variability, user customization, safety assurance, and precision path-planning as persistent challenges in robotic haircutting~\cite{khan2026cncinspired}. In practice, the problem is difficult not only because of motion accuracy, but also because tool use, constrained visual feedback, communication delay, and fault handling all become part of the control loop. Before autonomous haircutting can be evaluated responsibly, teleoperation studies are needed to characterize how sensing, communication, execution, tool actuation, and interruption handling interact during robotic haircutting.

For this reason, teleoperation provides a practical starting point for robotic haircutting research. It keeps the human operator in the loop and exposes system behavior for quantitative analysis. It also allows communication, execution, and operational safeguard behavior to be studied under controlled conditions before stronger autonomy is considered. Prior work provides useful pieces of this problem, but not yet the full system picture considered here. Related teleoperation studies have already shown the importance of communication-aware system behavior under deployed conditions~\cite{casavola2006predictive,chopra2008unreliable,beerens2020effect}. Related robotic studies near the head have focused more on grooming, hair interaction, perception, and task-level constraints than on dedicated haircutting systems. Recent work has also begun to frame haircutting robots from a mobile-robotics and deployment perspective~\cite{huang2026mobilehaircutting}, but accessible teleoperation studies have tended to emphasize navigation or general manipulation rather than near-head tool use. As a result, deployable teleoperation systems for robotic haircutting remain rarely reported, especially those combining monocular input, clipper actuation, remote deployment, and system-level analysis.

This paper presents TeleHairing, a prototype teleoperation system for robotic haircutting. The purpose is not to demonstrate autonomous haircutting or to assess haircut quality, but to study how a robotic haircutting system behaves under deployment, sensing, execution, tool-actuation, and interruption constraints. The paper makes three contributions. First, it introduces a task-specific teleoperation architecture for robotic haircutting that treats operator-side sensing, message transport, robot-side execution, clipper actuation, video feedback, and safeguards as separable but coordinated parts of one teleoperation loop. This modular design enables deployment-dependent communication effects, robot-side execution timing, and tool-state handling to be independently examined without changing the underlying haircutting task interface. Second, it establishes a repeatable mannequin-based evaluation framework for studying system-level robotic haircutting teleoperation under local, relay, and remote deployment conditions. Third, it quantitatively characterizes communication timing, command-following behavior, and interruption handling, thereby showing how the main components of the closed-loop teleoperation system can be decomposed and interpreted. Together, these contributions position TeleHairing as a quantitative reference for future research on robotic haircutting teleoperation.

The remainder of this paper is organized as follows. Section~\ref{sec.2} reviews related work on teleoperation, head-proximate grooming and haircutting systems, accessible deployable interfaces, and fault-handling mechanisms for \mbox{human-centered} robot operation. Section~\ref{sec.3} presents the TeleHairing architecture, including the communication layer, relative-pose control pipeline, tool actuation, and operational safeguards. Section~\ref{sec.4} describes the experimental setup and evaluation, focusing on deployment-dependent timing, proxy-task trajectory following, operator and network-path variation, and runtime fault handling. Section~\ref{sec.5} discusses the system-design implications, and Section~\ref{sec.6} concludes the paper.

\section{Background and Related Work}\label{sec.2}

Robotic haircutting is not only a motion-generation problem. It also requires an operator or autonomy stack to coordinate visual observation, communication, robot execution, tool actuation, and operational constraints around the head. The related work most relevant to TeleHairing therefore spans teleoperation, near-head hair manipulation, deployable service-robot interfaces, and fault-aware human-robot interaction. Each area provides part of the required capability, but none directly replaces a robotic haircutting teleoperation system that combines near-head tool motion, clipper actuation, deployment-dependent communication, and interruption handling within the same evaluated loop.

Teleoperation provides the first building block because it treats robotic manipulation as a closed loop among operator input, communication, remote execution, and sensory feedback~\cite{hokayem2006bilateral}. For remote robot operation over nonideal links, predictive teleoperation and networked-control studies have treated communication delay, quality-of-service variation, and controller-side timing as part of the control problem rather than as external implementation details~\cite{kruszewski2012switched}. Recent vision-based systems extend this closed-loop view to dexterous manipulation, demonstration collection, and immersive bimanual operation~\cite{qin2023anyteleop,cheng2024opentelevision}. Prior networked teleoperation studies, already discussed in the Introduction, further show that communication structure can affect remote manipulation. For systems intended to be operated across local and remote settings, performance-evaluation frameworks therefore motivate reporting latency, usability, and task behavior as system properties~\cite{wan2023performance}. However, this literature typically evaluates general remote manipulation, navigation, or dexterous tasks; it does not specify how the teleoperation loop changes when the robot must operate an actuated haircutting tool near the head.
\begin{figure*}[t]
	\centering
	\includegraphics[width=0.95\linewidth]{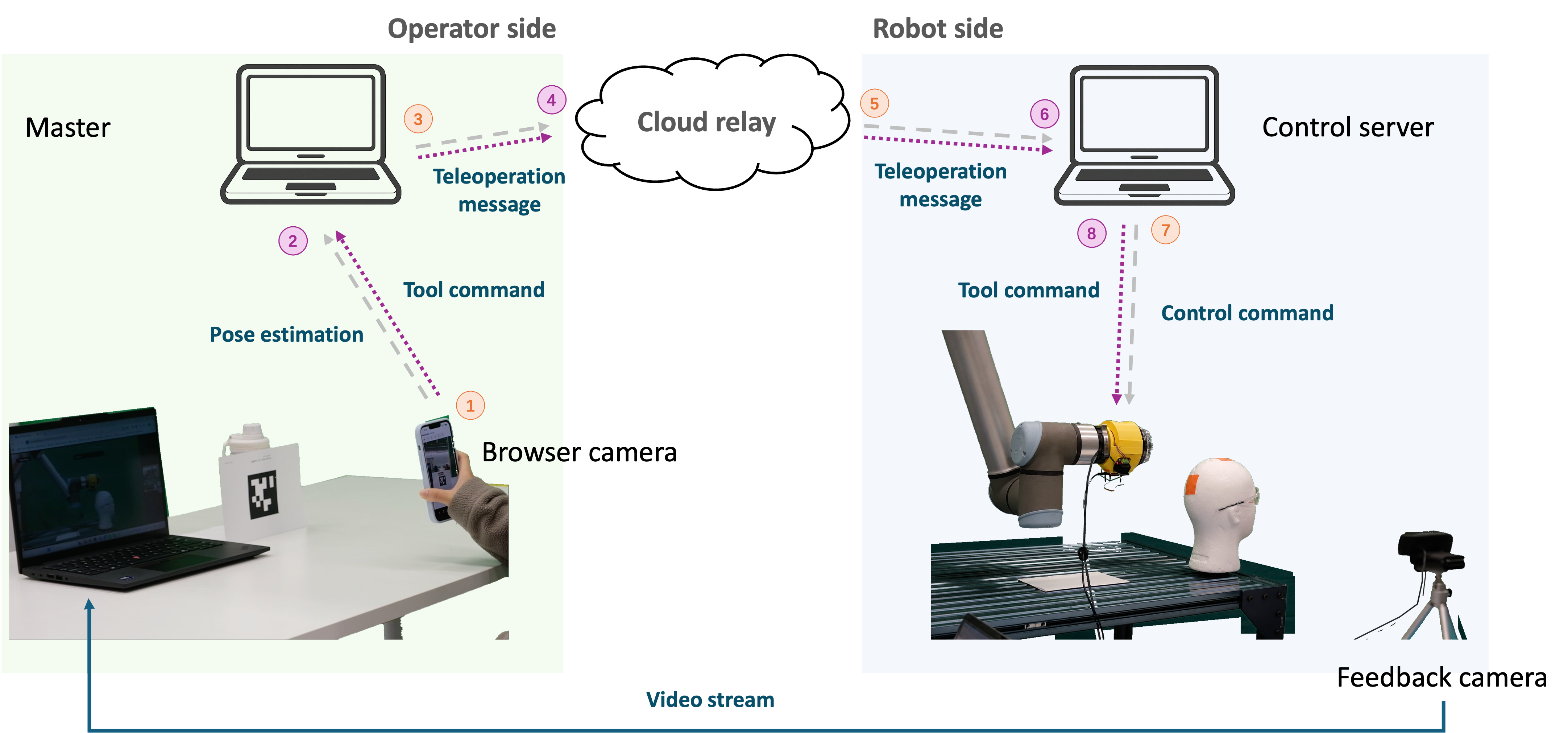}
	\caption{Architecture overview of TeleHairing. On the operator side, a phone camera supports pose estimation, and teleoperation messages are routed either directly or through a cloud relay to the robot-side control server. The control server coordinates manipulator and tool commands, while a feedback camera streams video back to the browser client to close the teleoperation loop. Orange numbered markers denote send-side or stage-start timestamps, whereas purple numbered markers denote receive-side or stage-end timestamps. The numbered pairs identify the timestamped timing segments reported in Table~\ref{tab:telehairing_latency}.}
	\label{fig:telehairing_arch}
\end{figure*}

The task-side constraints come from robotic hair interaction and near-head personal care. Existing haircutting reviews and the grooming systems cited in the Introduction identify perception, hair variability, contact-rich interaction, user preference, precision planning, and safety as persistent difficulties. Additional physical-platform studies reinforce this point from adjacent tasks: feedback-driven brushing shows how hair interaction can require online correction, while root-centric front-hair styling shows how robot motion can be planned around local hair geometry~\cite{hughes2021detangling,kim2025front}. These studies show that robotic interaction with hair near the head is feasible. Their focus, however, is mainly autonomous or semi-autonomous grooming, offline planning, or task-specific manipulation policies rather than deployable teleoperation for haircutting. They therefore do not characterize the combined behavior of human operation, communication, robot-side execution, and cutting-tool actuation during robotic haircutting teleoperation.

This connection raises a deployability question. A robotic haircutting teleoperation system should not rely only on a tightly integrated laboratory interface if the goal is to study local, relay, and remote operation. Accessible teleoperation studies of assistive mobile manipulators highlight the hardware, interface, and deployment constraints that arise when remote operators interact with robots in less controlled environments~\cite{cabrera2021exploration}. Visual fiducials such as AprilTags are widely used for robot pose estimation and calibration~\cite{olson2011apriltag}, and they offer a practical route to low-barrier operator-side visual input. These methods support deployable interfaces and remote operation, but they are commonly validated on navigation, fetch-and-deliver, or generic manipulation tasks. As a result, they do not establish how such lightweight operator-input methods, cloud-mediated control, robot-side execution, and video feedback perform when applied to clipper-based robotic haircutting near the head.

Near-head tool use also makes safety and fault handling part of robotic haircutting system behavior rather than an external concern. Prior work on virtual fixtures, steady-hand teleoperation, adaptive assistance, and fail-safe contact behavior shows how protective constraints can shape human-supervised robot motion near the body~\cite{rosenberg1993virtual,abbott2003steadyhand,aarno2005adaptivevf,yamada1996failsafe}. Haircutting-specific safety discussions further clarify that risks arise from tool motion near the scalp and from degraded sensing or control conditions~\cite{huang2026safetyhaircutting}. These studies provide important mechanisms and safety concepts, but they do not integrate tracking interruption, communication behavior, tool actuation, and robot recovery into one robotic haircutting teleoperation loop.

In summary, the missing element is not a single component such as a fiducial tracker, a teleoperation link, or a protective mechanism. The gap is a specifically designed and experimentally characterized robotic haircutting teleoperation system that brings these elements together. TeleHairing addresses this integration gap by studying a tool-actuated, locally and remotely deployable platform with logged timing, trajectory, and runtime fault-handling behavior.

\section{TeleHairing System Architecture}\label{sec.3}

This section presents the TeleHairing architecture and the design rationale underlying its use for robotic haircutting teleoperation. It describes how the operator side, optional cloud relay, and robot side form a closed loop, followed by the communication pathways, pose-based control method, clipper actuation design, and operational safeguards used for near-head operation.

\subsection{System Overview}\label{sec.3.1}

TeleHairing treats robotic haircutting as a supervised teleoperation task with near-head tool operation. In this task, the operator adjusts end-effector motion around the head from visual feedback while coordinating arm motion with haircutting-tool actuation. Because the system is intended for remote deployment, the communication link is treated as part of the task loop rather than as an external condition. Figure~\ref{fig:telehairing_arch} shows the overall architecture of TeleHairing. The architecture separates operator-side sensing, message transport, robot-side execution, tool actuation, and video feedback so that each element can be examined without changing the whole teleoperation loop.

At the architectural level, operator-side visual observation is represented as pose input relative to the initialized state and transmitted through either a direct or relay-mediated path. The robot side then applies the corresponding manipulator and tool commands, while an independent feedback camera closes the teleoperation control cycle.

This architecture serves three design purposes. First, separating operator-side sensing from message transport keeps visual processing local while allowing the transport path to change between local, relay, and remote deployments. Second, separating manipulator execution from clipper actuation keeps motion control and tool state as distinct robot-side processes. Third, separating the control-message path from the video-feedback path allows feedback bandwidth and control-message timing to be analyzed without conflating their communication structures.

\subsection{Communication Transport}\label{sec.3.2}

\textbf{\textit{Communication paths.}}
The design keeps visual pose extraction on the operator side while allowing the control transport path to vary across deployments. The operator-side phone camera observes the task area, including the AprilTag used as the teleoperation reference, and tag detection and pose estimation are performed locally in the browser client. This browser-side pose estimation decouples visual processing from wide-area transport: the networked control path carries pose-derived commands rather than raw video for remote processing.

In the current implementation, the real-time pose channel between the browser front end and the local client service uses WebRTC, with WebSocket fallback. The local client service is referred to as the master process in the timing logs. From this process, teleoperation messages are routed either through a direct local path to the robot-side control server or through a relay-mediated path via a cloud relay. This separation allows the browser-side interaction mechanism and the cross-network transport mechanism to be varied independently, which supports comparison of local, relay, and remote deployment conditions under the same robot-side execution logic.

For the evaluation in this paper, the same communication architecture is instantiated in three deployment modes:
\begin{itemize}
	\item \textbf{Local mode}: the operator side and the robot side communicate through a local-area network without a cloud relay. This mode provides the shortest-path reference condition.
	\item \textbf{Relay mode}: the operator side, cloud relay, and robot side all operate locally in Finland while still communicating through the cloud relay. This mode isolates the effect of relay insertion without adding an intercontinental path.
	\item \textbf{Remote mode}: the operator is located in Hong Kong, while the cloud relay and robot side remain in Finland. This mode records cross-region deployment behavior under a remote-operator condition.
\end{itemize}

The control-message path combines pose-related commands and tool-switching commands within one session context, while robot-side processing separates manipulator execution from tool control. This design keeps the operator command stream coherent while preserving a logical distinction between spatial motion and clipper state. In parallel, the feedback camera streams task video back to the browser client through an independent feedback path. Keeping video feedback outside the control-message path prevents feedback bandwidth and streaming behavior from directly changing the control-message architecture.

\textbf{\textit{Timing and synchronization.}}
The timing analysis follows the software stages of the teleoperation path. On the operator side, the browser records the tag-detection and command-transmission instants. The master process records message reception and forwarding. In relay-based modes, the cloud relay records message arrival and forwarding, and the robot-side control server records socket reception, control-loop processing, command issue, and the next robot-state sample obtained through the Real-Time Data Exchange (RTDE) interface. Thus, the communication metrics reported in Section~\ref{sec.4.2} are software timestamp differences across the teleoperation chain, rather than hardware-instrumented oscilloscope measurements.

Before each capture, active timing links were synchronized by an application-level two-way exchange. This was not a manual correction and was not treated as an external NTP-calibrated measurement. For each active hop, 20 round-trip samples were collected and used to estimate the clock offset between the two software endpoints under the usual symmetric-delay assumption of two-way timing exchange. The corresponding round-trip times (RTTs) are reported with Table~\ref{tab:telehairing_latency} because they indicate the practical uncertainty scale for one-way cross-node latency estimates. Processing intervals measured inside one process are less sensitive to clock offset, whereas one-way links between different hosts can be affected by residual clock-offset error, path asymmetry, and network jitter. Therefore, small differences close to this scale should be interpreted conservatively.

For the robotic haircutting setting considered here, this communication structure yields two system properties used later in the evaluation. First, pose input, tool switching, and execution control share a unified session context, which supports consistent command interpretation across direct and relay-based deployments. Second, video feedback remains architecturally separate from control-message transport, so deployment-dependent timing can be analyzed on the control path without treating video streaming as part of the command chain.

\subsection{Pose-Based Teleoperation Control}\label{sec.3.3}

TeleHairing uses a single AprilTag~\cite{olson2011apriltag} as the visual reference for teleoperation. On the operator-input side, two coordinate frames are used at this stage: the tag frame $\{\mathrm{tag}\}$ and the camera frame $\{\mathrm{cam}\}$. At time $t$, the client provides the camera pose expressed in the tag frame,
\begin{equation}
	{}^{\mathrm{tag}}\mathbf{T}_{\mathrm{cam},t}
	=
	\begin{bmatrix}
		{}^{\mathrm{tag}}\mathbf{R}_{\mathrm{cam},t} & {}^{\mathrm{tag}}\mathbf{p}_{\mathrm{cam},t} \\
		\mathbf{0}^{\top} & 1
	\end{bmatrix},
\end{equation}
where ${}^{\mathrm{tag}}\mathbf{T}_{\mathrm{cam},t}$ denotes the camera pose at time $t$, expressed in the tag frame. The initialized pose is written compactly as ${}^{\mathrm{tag}}\mathbf{T}_{\mathrm{cam},0}$. The left superscript identifies the reference frame in which the quantity is expressed, and the subscript identifies the observed frame. In the homogeneous transform, $\mathbf{R}\in\mathrm{SO}(3)$ is a rotation matrix and $\mathbf{p}\in\mathbb{R}^{3}$ is a translation vector.

The robot side does not use this input representation directly. Instead, the current pose and the initialized pose are first mapped into a robot-side control convention. Let this fixed mapping be denoted by $\mathcal{M}(\cdot)$. In the current implementation, $\mathcal{M}(\cdot)$ represents a fixed axis and sign convention conversion from the browser-side pose output to the robot-side control convention, rather than an online hand-eye calibration estimate. The mapped current pose is written as
\begin{equation}
	\mathbf{T}_{t}^{\mathrm{ctrl}}
	=
	\mathcal{M}\!\left({}^{\mathrm{tag}}\mathbf{T}_{\mathrm{cam},t}\right)
	=
	\begin{bmatrix}
		\mathbf{R}_{t}^{\mathrm{ctrl}} & \mathbf{p}_{t}^{\mathrm{ctrl}} \\
		\mathbf{0}^{\top} & 1
	\end{bmatrix},
\end{equation}
where $\mathbf{T}_{t}^{\mathrm{ctrl}}$ denotes the mapped input pose at time $t$ under the robot-side control convention. The corresponding initialized pose is written as $\mathbf{T}_{0}^{\mathrm{ctrl}}$, with rotational and translational parts denoted by $\mathbf{R}_{0}^{\mathrm{ctrl}}$ and $\mathbf{p}_{0}^{\mathrm{ctrl}}$, respectively. The translational and rotational increments relative to this initialized state are then defined as
\begin{equation}
	\Delta \mathbf{p}_{t}^{\mathrm{ctrl}}
	=
	\mathbf{p}_{t}^{\mathrm{ctrl}}
	-\mathbf{p}_{0}^{\mathrm{ctrl}},
	\qquad
	\Delta \mathbf{R}_{t}^{\mathrm{ctrl}}
	=
	\mathbf{R}_{t}^{\mathrm{ctrl}}
	\left(\mathbf{R}_{0}^{\mathrm{ctrl}}\right)^{\top}.
\end{equation}
where $\Delta \mathbf{p}_{t}^{\mathrm{ctrl}}$ is the translational displacement and $\Delta \mathbf{R}_{t}^{\mathrm{ctrl}}$ is the rotational change, both relative to the initialized state.

Let the initialized tool-center-point (TCP) pose be denoted by $\mathbf{T}_{\mathrm{tcp},0}$, with rotational and translational parts $\mathbf{R}_{\mathrm{tcp},0}$ and $\mathbf{p}_{\mathrm{tcp},0}$, respectively. This initialized pose is obtained from the robot through the RTDE interface. At time $t$, the desired TCP position $\mathbf{p}_{\mathrm{tcp},t}^{\mathrm{des}}$ and orientation  $\mathbf{R}_{\mathrm{tcp},t}^{\mathrm{des}}$ are then generated by applying these increments to the initialized TCP pose:
\begin{equation}
	\mathbf{p}_{\mathrm{tcp},t}^{\mathrm{des}}
	=
	\mathbf{p}_{\mathrm{tcp},0}
	+
	\Delta \mathbf{p}_{t}^{\mathrm{ctrl}},
	\qquad
	\mathbf{R}_{\mathrm{tcp},t}^{\mathrm{des}}
	=
	\Delta \mathbf{R}_{t}^{\mathrm{ctrl}}
	\mathbf{R}_{\mathrm{tcp},0}.
\end{equation}

In summary, TeleHairing first maps the input pose into the robot-side control convention, then computes translational and rotational increments relative to the initialized state, and finally applies those increments to the initialized TCP pose.

This pose-based control path depends on monocular AprilTag pose estimation. Its accuracy can be affected by the tag size, the 640 $\times$ 480 camera stream, viewing angle, lighting, motion blur, and camera calibration. The filtering, deadband, and speed-limit steps reduce high-frequency jitter before execution, but they do not remove all sensing error. In the present study, AprilTag pose-estimation accuracy was not independently characterized; it is treated as part of the sensing condition for the command-following analysis.

\subsection{Clipper Actuation Design}\label{sec.3.4}

\begin{figure}[!t]
	\centering
	\includegraphics[width=0.5\linewidth]{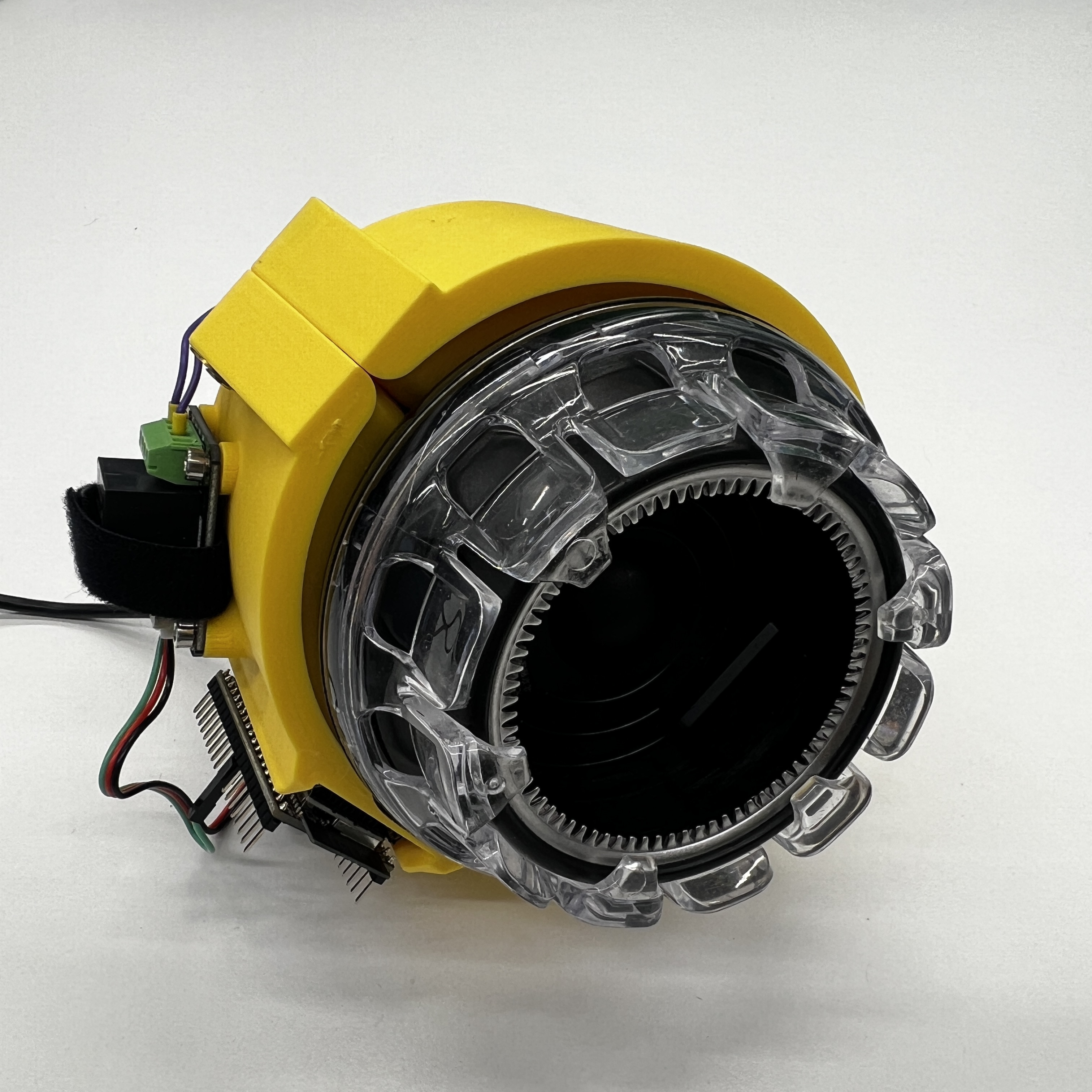}
	\caption{Robot-side clipper integration used in TeleHairing.}
	\label{fig:telehairing_clipper}
\end{figure}

TeleHairing uses a commercial hair clipper as the robot-mounted end effector, as shown in Figure~\ref{fig:telehairing_clipper}. The clipper is fixed to the robot flange through a dedicated 3D-printed adapter, which preserves a repeatable mounting pose across trials. Its original manual switching circuit was modified into an electrically triggered interface by connecting the clipper circuit to a relay controlled by an STM32 board. During teleoperation, binary on/off commands are issued from the browser client and applied on the robot side through relay switching. This design makes the clipper state part of the logged teleoperation loop while keeping it logically separate from Cartesian manipulator execution. This separation preserves simple binary tool actuation, avoids embedding tool state inside the motion controller, and allows tool commands to be handled alongside robot-motion commands while remaining logically distinct.

\subsection{Operational Safeguards}\label{sec.3.5}

TeleHairing implements three operational safeguards in the server-side control loop.

\begin{itemize}
	\item \textbf{Speed limitation.} Position and rotation increments are subjected to velocity-based limiting before they enter the downstream control stages. The update interval $\Delta t$ is not fixed. It is determined from the time difference between consecutive valid pose events. Because these events come from tag detections in the camera stream, $\Delta t$ is directly related to the camera frame rate and the rate of valid detection output. When the camera runs at a nominal frame rate $f_{\mathrm{cam}}$ and each frame yields a valid detection, the interval is approximately
	\begin{equation}
		\Delta t \approx \frac{1}{f_{\mathrm{cam}}}.
	\end{equation}
	If detections are missed, frames are dropped, or output intervals become longer, then $\Delta t$ increases accordingly. Let $v_{\max}$ and $\omega_{\max}$ denote the maximum translational and angular velocities. The corresponding maximum translational and rotational steps for a single update are
	\begin{equation}
		\Delta p_{\max} = v_{\max}\,\Delta t,
		\qquad
		\Delta \theta_{\max} = \omega_{\max}\,\Delta t.
	\end{equation}
	If the translational increment norm $\lVert \Delta \mathbf{p}_{t} \rVert$ or the rotational increment angle $\Delta \theta_{t}$ exceeds these limits, the system scales the increment proportionally so that
	\begin{equation}
		\lVert \Delta \mathbf{p}_{t}^{\,\mathrm{lim}} \rVert \leq \Delta p_{\max},
		\qquad
		\Delta \theta_{t}^{\,\mathrm{lim}} \leq \Delta \theta_{\max}.
	\end{equation}
	When a valid event interval is unavailable, the implementation falls back to the nominal control period as an approximation of $\Delta t$. This limits sudden motion under large operator inputs, frame-rate variation, detection interruption, or irregular event spacing.
	
	\item \textbf{Rebase.} When the tag temporarily leaves the field of view, the system does not simply reconnect the re-detected pose to the previous control chain. Instead, it performs a rebase step. Once the tracking gap reaches the configured threshold, the newly reacquired tag pose becomes the new initial reference pose and the current maintained robot target becomes the new robot anchor. This allows control to resume from the current robot state after reacquisition without introducing a discontinuous target jump.
	
	\item \textbf{Protective stop recovery.} In the current implementation, the control session is terminated when the UR controller enters a protective stop and streamed robot control can no longer continue. The server treats this event as a stop condition for the session. Recovery is not automatic. The operator must manually clear the protective stop and recover the robot from the teach pendant before the system can return to normal operation.
\end{itemize}

Overall, TeleHairing links operator-side pose sensing, direct or relay-mediated message transport, robot-side manipulator and tool execution, and independent video feedback within one logged teleoperation loop. This structure supports the experimental analysis of communication behavior, trajectory behavior, and runtime fault handling.

\begin{table*}[!t]
	\centering
	\caption{System and experimental configuration used in the TeleHairing evaluation.}
	\label{tab:telehairing_setup_config}
	{\footnotesize
		\setlength{\tabcolsep}{5pt}
		\begin{tabular}{p{4cm} p{12cm}}
			\toprule
			\textbf{Item} & \textbf{Configuration} \\
			\midrule
			Robot platform & UR10e 6-degree-of-freedom manipulator \\
			End effector & Commercial hair clipper mounted through a dedicated 3D-printed adapter and electrically switched through a relay and STM32 interface \\
			Control command interface & RTDE-based Cartesian command streaming \\
			Controller frequency & 30~Hz robot-side Cartesian command updates \\
			Velocity limits & 0.05~m/s translational limit and 15~deg/s rotational limit \\
			Visual reference & Single AprilTag reference marker: tag36h11, ID~0, physical tag size 75~mm \\
			Operator-side camera & Rear smartphone camera, with monocular pose estimation computed in the browser client from a 640 $\times$ 480 video stream at 30~fps \\
			Feedback camera & Robot-side task-scene video camera for closed-loop observation \\
			Cloud relay deployment & Relay server deployed on CSC Pouta cloud infrastructure in Kajaani, Finland \\
			Communication stack & Browser-side real-time pose channel preferring WebRTC with WebSocket fallback, direct local control forwarding or cross-network forwarding through a cloud relay, and a video-feedback path separate from the control-message path \\
			Clock synchronization & Hop-wise initialization synchronization over each active timing link, using 20 round-trip samples per link; active-hop RTT summaries are reported with Table~\ref{tab:telehairing_latency} \\
			Deployment modes & Local mode, relay mode, and remote mode \\
			Normal-trial counts & 10 local-mode trials, 10 relay-mode trials, and 10 remote-mode trials \\
			Fault-condition counts & 5 detection-loss trials and 1 protective-stop recovery example \\
			Task trajectory & Mannequin-based proxy trajectory for robotic haircutting evaluation \\
			\bottomrule
		\end{tabular}
	}
\end{table*}

\begin{table*}[!t]
	\centering
	\caption{Stage-wise communication timing across the three TeleHairing modes. Values are reported in milliseconds.}
	\label{tab:telehairing_latency}
	{\footnotesize
		\setlength{\tabcolsep}{3pt}
		\begin{tabular}{p{2.5cm} >{\centering\arraybackslash}p{2.5cm} p{2.7cm} p{2.7cm} p{2.7cm}}
			\toprule
			\textbf{Timing segment} & \textbf{Timestamp pair} & \textbf{Local mode} & \textbf{Relay mode} & \textbf{Remote mode} \\
			\midrule
			Browser$\to$master      & 1--2     & 7.4 $\pm$ 2.2   & 9.3 $\pm$ 4.0   & 1.3 $\pm$ 3.8 \\
			Master proc             & 2--3     & 0.1 $\pm$ 0.3   & 0.1 $\pm$ 0.6   & 0.2 $\pm$ 0.4 \\
			Master$\to$cloud        & 3--4     & ---             & 14.4 $\pm$ 2.7  & 116.9 $\pm$ 2.0 \\
			Cloud proc              & 4--5     & ---             & 0.0 $\pm$ 0.2   & 0.0 $\pm$ 0.2 \\
			Cloud$\to$control       & 5--6     & ---             & 5.5 $\pm$ 2.2   & 5.7 $\pm$ 4.0 \\
			Master$\to$control      & 3--6     & 1.1 $\pm$ 0.8   & ---             & --- \\
			Command queue            & internal & 0.1 $\pm$ 0.2   & 0.1 $\pm$ 0.3   & 0.1 $\pm$ 0.3 \\
			Control proc            & 6--7     & 19.4 $\pm$ 10.9 & 18.9 $\pm$ 10.6 & 19.1 $\pm$ 10.6 \\
			Control$\leftrightarrow$robot & 7--8     & 35.8 $\pm$ 1.8  & 35.6 $\pm$ 1.9  & 35.6 $\pm$ 1.8 \\
			Entire time             & 1--8     & 75.6 $\pm$ 11.2 & 95.7 $\pm$ 11.3 & 190.5 $\pm$ 11.4 \\
			\bottomrule
		\end{tabular}
	}
	
	\vspace{2pt}
	\begin{minipage}{\linewidth}
		\footnotesize
		\textit{Remark.} Timestamp pairs refer to the numbered markers in Figure~\ref{fig:telehairing_arch}. ``Internal'' denotes the control-server command queue; ``---'' denotes a topology-dependent non-applicable segment. Clock-sync RTTs are reported as mean $\pm$ SD in ms: local, browser--master 10.60 $\pm$ 0.97 and master--control 0.61 $\pm$ 0.09; relay, browser--master 10.40 $\pm$ 3.34, master--cloud 12.50 $\pm$ 1.28, and cloud--control 8.36 $\pm$ 0.28; remote, browser--master 6.30 $\pm$ 0.48, master--cloud 227.04 $\pm$ 0.18, and cloud--control 8.30 $\pm$ 0.36.
	\end{minipage}
\end{table*}

\section{Experimental Evaluation}\label{sec.4}

The experiments evaluate the teleoperation performance of TeleHairing for robotic haircutting. The present study focuses on system-level behavior rather than haircut quality and therefore employs a repeatable mannequin-based proxy trajectory under local, relay, and remote deployment conditions. The dependent variables are communication timing across software endpoints, command-following trajectory error, and runtime fault-handling behavior. Within this scope, the analysis reports the experimental configuration, deployment-dependent timing behavior, trajectory-following behavior under the mannequin-based proxy task, and runtime responses to representative interruptions.

\subsection{Experimental Setup}\label{sec.4.1}

All reported trials shared the same core robotic haircutting teleoperation configuration: robot platform, clipper-equipped end effector, mannequin-head proxy task, and robot-side Cartesian command update frequency of 30~Hz. Robot-side command updates were velocity-limited, with a translational limit of 0.05~m/s and a rotational limit of 15~deg/s. On the operator side, a monocular rear smartphone camera and a browser-based client generated relative pose commands from a single AprilTag reference. On the robot side, the mapped Cartesian commands were executed by the control server, which also logged robot motion, timing, and runtime fault-handling events.

Table~\ref{tab:telehairing_setup_config} summarizes the fixed system and experiment configuration used in the present evaluation. Within this setup, the fixed conditions were the robot platform, end effector, control frequency, velocity limits, monocular AprilTag input method, proxy trajectory definition, initialization procedure, and robot-side execution and logging logic. The main experimental variable was the deployment mode. Operator identity was not treated as an independent study factor. However, the remote-mode trials were performed by a different operator, so operator identity remains a standing confound for trajectory comparison between remote mode and the two Finland-based modes.

The three deployment modes used in the experiment follow the definitions given in Section~\ref{sec.3.2}. The local-mode and relay-mode trials were performed by the same local operator, whereas the remote-mode trials were performed by a different remote operator. Before each trial, the system was initialized through time synchronization and initial calibration, followed by reference registration for the fiducial-based control loop. After each stop event, the next trial was restarted from the same standard initialization sequence and start condition.

The mannequin-based proxy trajectory was used to evaluate teleoperation behavior in a repeatable robotic haircutting setting rather than to assess hairstyle outcome or cutting quality. In each normal trial, the operator guided the clipper-equipped tool through a fixed path around the mannequin head, moving from the posterior head region toward the crown and back to the initial region before returning to the nominal start condition. This proxy trajectory was designed to test whether the teleoperation system could maintain continuous near-head tool motion with monocular input across the three evaluated deployment conditions. The dataset contains 36 captures collected under this protocol: 10 local-mode normal trials, 10 relay-mode normal trials, 10 remote-mode normal trials, 5 detection-loss trials, and 1 protective-stop recovery example. These data support system-level analysis of robotic haircutting teleoperation behavior under the tested conditions, but not broad claims about haircut quality, long-term robustness, operator generality, or human-use readiness.

\begin{figure*}[t]
	\centering
	\subfloat[{\scriptsize Local mode\label{fig:telehairing_trajectory_local}}]{\includegraphics[width=0.32\linewidth]{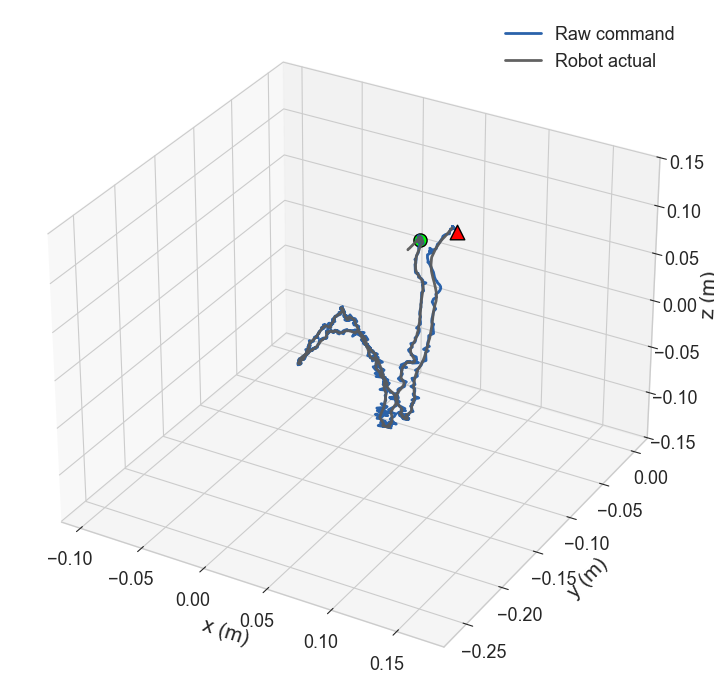}}
	\hfil
	\subfloat[{\scriptsize Relay mode\label{fig:telehairing_trajectory_cloud_fin}}]{\includegraphics[width=0.32\linewidth]{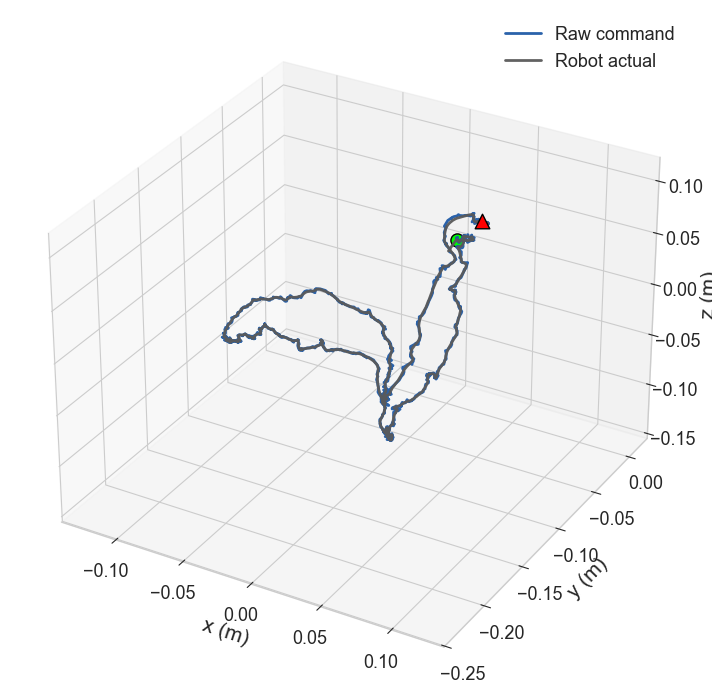}}
	\hfil
	\subfloat[{\scriptsize Remote mode\label{fig:telehairing_trajectory_cloud_hk}}]{\includegraphics[width=0.32\linewidth]{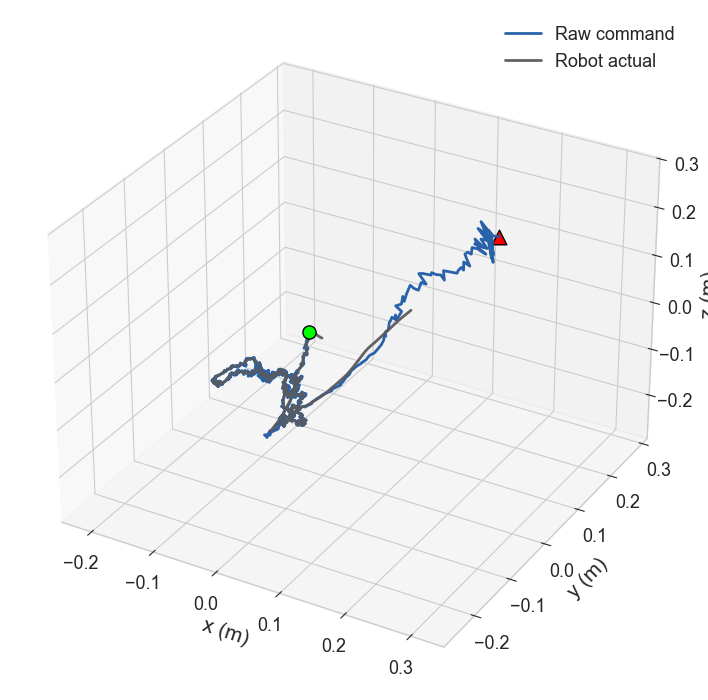}}
	\caption{Representative command and measured trajectory overlays for the three TeleHairing modes.}
	\label{fig:telehairing_trajectory_overlays}
\end{figure*}

\begin{table*}[t]
	\centering
	\caption{Command-following trajectory error across the three TeleHairing modes.}
	\label{tab:telehairing_trajectory_summary}
	{\footnotesize\setlength{\tabcolsep}{4pt}
		\begin{tabular}{p{4.2cm} p{1.0cm} p{3.5cm} p{3.5cm}}
			\toprule
			\textbf{Mode} & \textbf{N} & \textbf{RMSE (mm)} & \textbf{Peak error (mm)} \\
			\midrule
			Local mode               & 10 & 6.7 $\pm$ 1.7   & 27.3 $\pm$ 12.9  \\
			Relay mode               & 10 & 5.1 $\pm$ 0.9   & 19.6 $\pm$ 6.5   \\
			Remote mode              & 10 & 45.1 $\pm$ 17.1 & 223.6 $\pm$ 80.8 \\
			Remote mode (first 90\%) & 10 & 9.6 $\pm$ 3.0   & 34.9 $\pm$ 10.5  \\
			\bottomrule
	\end{tabular}}
	
	\vspace{2pt}
	\begin{minipage}{\linewidth}
		\footnotesize
		\textit{Remark.} Values are mean $\pm$ SD in millimeters across trials. RMSE and peak error are computed per trial from valid command--TCP sample pairs. The full-trajectory remote-mode result remains the primary outcome; the first-90\% row is only a diagnostic analysis that excludes the terminal rapid-withdrawal segment in the remote trials.
	\end{minipage}
\end{table*}
\subsection{Communication Performance}\label{sec.4.2}

Deployment-dependent timing changes mainly occur before commands reach the robot-side control endpoint, while robot-side queueing, control processing, and control-to-robot timing remain largely stable across modes. The stage-wise timing decomposition in Table~\ref{tab:telehairing_latency} supports this observation, with each row tied to a logged timestamp pair marked in Figure~\ref{fig:telehairing_arch}. Because the timing decomposition combines timestamps generated on different processes and hosts, the clock-sync RTT values reported with Table~\ref{tab:telehairing_latency} provide the main measurement context for interpreting cross-node latency segments. The large remote-mode increase in Master$\to$cloud is well separated from this synchronization variability.

The stage-wise decomposition supports this interpretation. Between local mode and relay mode, Browser$\to$master and Master proc remain similar, while relay mode adds 14.4~ms in Master$\to$cloud and 5.5~ms in Cloud$\to$control. In remote mode, the largest further increase is observed in Master$\to$cloud, which reaches 116.9~ms. Cloud$\to$control remains close to relay mode at 5.7~ms. By contrast, the robot-side timing remains similar across all three modes: Command queue remains at 0.1~ms, Control proc remains around 19~ms, and Control$\leftrightarrow$robot, measured from control-command issue to the next robot-state sample, remains around 35.6~ms.

These stage-wise differences explain the increase in the aggregate Entire time, which rises from 75.6~ms in local mode to 95.7~ms in relay mode and 190.5~ms in remote mode. Within the current timing breakdown, the main timing differences across the three modes arise before control messages reach the robot side.

\subsection{Trajectory-Following Behavior}\label{sec.4.3}

Trajectory-following behavior was compared across the three modes under the same mannequin-based proxy task and sensing setup. Figure~\ref{fig:telehairing_trajectory_overlays} visualizes the last recorded trial from each mode, whereas Table~\ref{tab:telehairing_trajectory_summary} reports the summary statistics across all 10 trials per mode.

For each valid teleoperation sample $i$, the translation error was computed as
\begin{equation}
	e_i =
	\left\|
	\Delta \mathbf{p}_{i}^{\mathrm{meas}}
	-
	\Delta \mathbf{p}_{i}^{\mathrm{cmd}}
	\right\|_2 ,
\end{equation}
where $\Delta \mathbf{p}_{i}^{\mathrm{cmd}}$ is the recorded relative translation command and $\Delta \mathbf{p}_{i}^{\mathrm{meas}}$ is the recorded robot TCP displacement from the start of the trial. The RMSE is $\sqrt{N^{-1}\sum_{i=1}^{N} e_i^2}$ over all valid samples, and the peak error is $\max_i e_i$. The command and TCP displacement were paired by the same teleoperation sample in the system log, without external time-resampled alignment or independent motion-capture reference. These values should therefore be interpreted as system-level command-following errors rather than absolute haircut-path errors, because they reflect both spatial tracking error and phase lag introduced by communication delay, command queueing, robot update timing, and velocity limiting.

\begin{figure*}[t]
	\centering
	\subfloat[\label{fig:telehairing_safeguard_timeline}]{\includegraphics[width=0.34\linewidth]{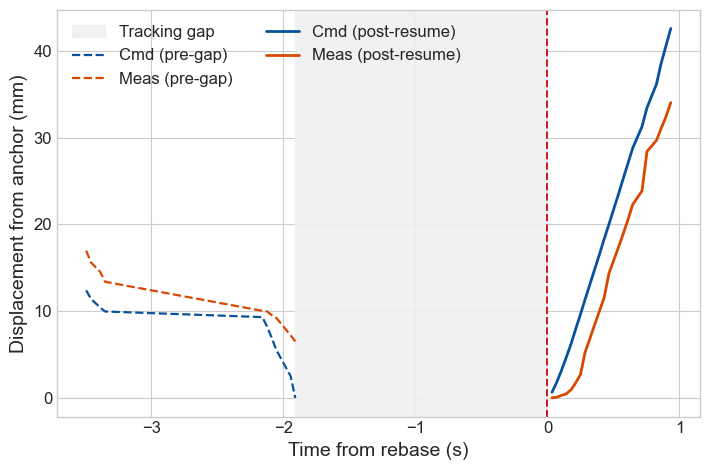}}
	\hfil
	\subfloat[\label{fig:telehairing_safeguard_resume}]{\includegraphics[width=0.31\linewidth]{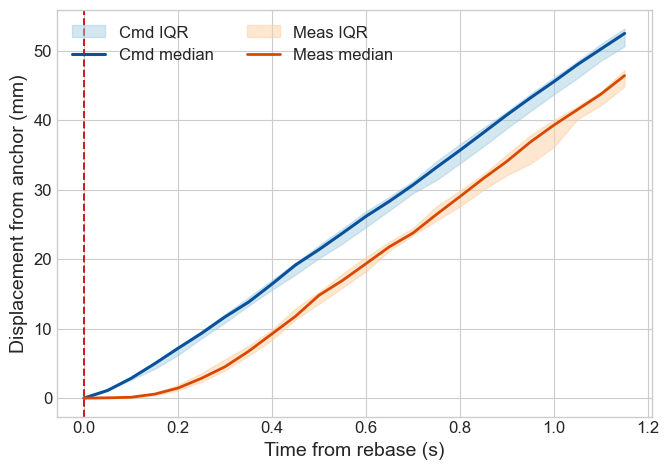}}
	\hfil
	\subfloat[\label{fig:telehairing_safeguard_drift}]{\includegraphics[width=0.31\linewidth]{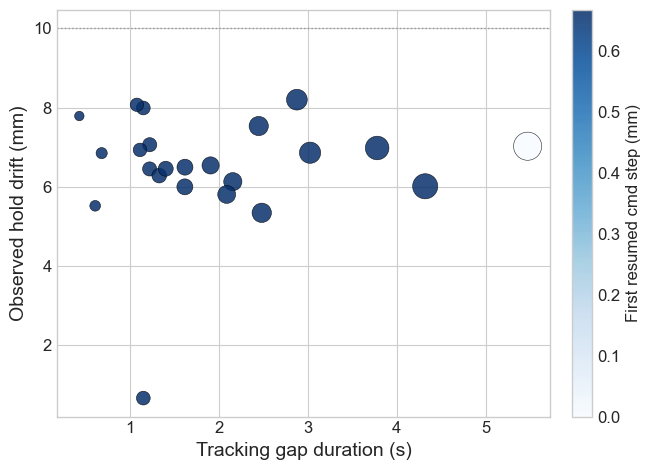}}
	\caption{Detection-loss recovery behavior in the dedicated trials. Panel (a) shows a single recovery event with the tracking gap aligned to the rebase instant; panel (b) shows the post-recovery median and interquartile displacement profile; and panel (c) shows the relation between tracking-gap duration, accumulated hold drift, and the first resumed command step.}
	\label{fig:telehairing_safeguard_overview}
\end{figure*}

The remote-mode command-following error is strongly concentrated in the terminal withdrawal segment rather than distributed throughout the trajectory. Figure~\ref{fig:telehairing_trajectory_overlays} shows that the local-mode and relay-mode examples remain close to the commanded proxy path over the full robotic haircutting trial, whereas the remote-mode example exhibits a visible following gap near the end of the trial. Table~\ref{tab:telehairing_trajectory_summary} quantifies the same pattern: the full remote-mode trajectory has a mean RMSE of 45.1~mm and a mean peak error of 223.6~mm, whereas retaining only the first 90\% of each remote-mode trajectory reduces these values to 9.6~mm and 34.9~mm, respectively. This indicates that the main remote trajectory deviation appears during the terminal withdrawal rather than as a persistent tracking offset over the full path.

This trajectory pattern should not be attributed to latency alone. Section~\ref{sec.4.2} shows that remote mode has substantially higher pre-control communication delay, dominated by the Master$\to$cloud segment, but the remote-mode trials were also performed by a different operator. In those trials, the operator rapidly withdrew the input after reaching the designated end region; this terminal command change interacted with the robot-side velocity limit, so the robot TCP had not yet reached the commanded state when the capture was stopped. The monocular AprilTag input may also contribute sensing error, although that sensing setup was held fixed across all three modes. The trajectory result should therefore be interpreted as behavior observed under the tested remote deployment condition, where increased pre-control delay, operator behavior, and velocity-limited execution were present together.

\subsection{Runtime Fault-Handling Behavior}\label{sec.4.4}

Runtime fault-handling behavior was evaluated under detection-loss and protective-stop interruptions. The analysis is based on five dedicated detection-loss trials and one protective-stop trial.

In the detection-loss trials, the fiducial was intentionally occluded during the mannequin-based proxy trajectory and then made visible again. Figure~\ref{fig:telehairing_safeguard_overview}(a)--(c) summarizes the observed interruption and recovery behavior, and Table~\ref{tab:telehairing_rebase_indicators} reports the corresponding quantitative indicators. Across the five trials, 23 re-acquisition and rebase events were recorded. During these events, the measured hold drift remained limited, and the first resumed command step was small, indicating that motion resumed from the re-anchored reference without a large command jump. These results characterize recovery behavior under the tested detection-loss condition, but they should not be interpreted as safety-boundary validation or a guarantee of near-head safety.

\begin{table}[t]
	\centering
	\caption{Quantitative indicators for detection-loss recovery.}
	\label{tab:telehairing_rebase_indicators}
	{\footnotesize
		\begin{tabular}{p{6cm} p{6cm}}
			\toprule
			\textbf{Indicator} & \textbf{Value} \\
			\midrule
			Recorded events & 23 events from 5 trials \\
			Effective tracking gap & 1.97 $\pm$ 1.25~s; range 0.43--5.47~s \\
			Missing updates & 53.61 $\pm$ 34.71 \\
			Measured hold drift & 6.47 $\pm$ 1.50~mm \\
			First resumed command step & 0.64 $\pm$ 0.14~mm \\
			\bottomrule
		\end{tabular}
	}
\end{table}

The protective-stop result is reported only as an operational observation because it is based on a single recorded trial. In that trial, the control stream stopped after the UR controller entered protective stop, and teleoperation resumed only after manual pendant recovery and re-initialization. Unlike tag-loss recovery, protective-stop recovery therefore required explicit operator intervention before the session could continue.

\section{Discussion}\label{sec.5}

The timing results indicate that wide-area deployment changes the teleoperation loop mainly before commands reach the robot-side control endpoint. In the tested remote condition, the largest latency increase appears in the Master$\to$cloud segment, whereas command queueing, control processing, and control-to-robot timing remain similar across deployment modes. For robotic haircutting system design, this distinction is important: remote deployment does not necessarily change the local robot execution process, but it changes when operator intent arrives at that process. This suggests that future systems should treat pre-control communication as a deployment-dependent layer while preserving stable robot-side execution and logging interfaces.

The trajectory results further show that increased communication delay alone is insufficient to explain remote teleoperation behavior. The remote-mode error is concentrated near the terminal withdrawal, where operator input changed rapidly and the velocity-limited robot-side execution could not immediately follow. This interaction matters for robotic haircutting because operators may make abrupt corrective or withdrawal motions near the end of a local cutting pass. Interface design and robot-side command handling should therefore consider not only average latency, but also how delayed operator intent interacts with speed limits, task termination, and near-head withdrawal behavior.

The detection-loss trials highlight interruption and re-entry as a separate design issue for robotic haircutting teleoperation. Rebase events allowed motion to resume from a re-anchored reference with limited hold drift and a small first resumed command step under the tested condition. This result suggests that future interfaces may benefit from treating interruption handling as an explicit part of the control loop rather than treating tracking loss as a simple communication dropout. In near-head tool operation, future studies should examine how re-entry interfaces preserve continuity of the operator reference, communicate the re-anchoring state to the operator, and avoid large resumed-command jumps.

\section{Conclusion}\label{sec.6}

This paper formulates robotic haircutting as a system-level teleoperation problem and presents TeleHairing as an experimentally characterized teleoperation baseline. The system combines monocular AprilTag pose input, local or \mbox{cloud-relay} communication, robot-side Cartesian execution, independent clipper switching, video feedback, and operational safeguards. The mannequin-based evaluation identifies three system-level behaviors relevant to robotic haircutting teleoperation: deployment-dependent timing changes occur mainly before the robot-side control endpoint, the larger remote trajectory error is concentrated in the terminal withdrawal segment, and detection-loss rebase events allow motion to resume without a large command jump under the tested condition. These results indicate how communication, operator input, velocity-limited execution, and interruption handling jointly shape the behavior of a near-head robotic haircutting loop. The present evaluation is limited to mannequin-based system-level teleoperation and does not assess haircut quality, human-subject operation, or formal safety validation. Within this scope, TeleHairing provides an experimentally characterized teleoperation foundation for future robotic haircutting systems.

\bibliographystyle{unsrt}  
\bibliography{references}

\end{document}